# Parameter-Efficient Fine-Tuning for Medical Text Summarization: A Comparative Study of LoRA, Prompt Tuning, and Full Fine-Tuning

Ulugbek Shernazarov [1], Rostislav Svitsov [1] and Bin Shi [1]

Telecom SudParis, Institut Polytechnique de Paris, France

*ABSTRACT*

*Fine-tuning large language models for domain-specific tasks such as medical text summarization demands substantial computational resources. Parameter-efficient fine-tuning (PEFT) methods offer promising alternatives by updating only a small fraction of parameters. This paper compares three adaptation approaches—Low-Rank Adaptation (LoRA), Prompt Tuning, and Full Fine-Tuning—across the Flan-T5 model family on the PubMed medical summarization dataset. Through experiments with multiple random seeds, we demonstrate that LoRA consistently outperforms full fine-tuning, achieving $43.52 \pm 0.18$ ROUGE-1 on Flan-T5-Large with only 0.6% trainable parameters compared to $40.67 \pm 0.21$ for full fine-tuning. Sensitivity analyses examine the impact of LoRA rank and prompt token count. Our findings suggest the low-rank constraint provides beneficial regularization, challenging assumptions about the necessity of full parameter updates.Code is available at https://github.com/eracoding/llm-medical-summarization*

## 1. Introduction

Medical text summarization has become increasingly important in healthcare informatics as clinicians and researchers face the challenge of processing an ever-growing volume of biomedical literature. PubMed alone indexes over 1.5 million new articles each year, making it practically impossible for healthcare professionals to stay current with developments in their fields through manual reading [1]. Automated summarization systems address this challenge by distilling lengthy scientific articles into concise summaries that preserve essential findings, thereby supporting evidence-based medicine and clinical decision-making.

The emergence of large language models (LLMs) has substantially advanced the state of the art in text summarization [2]. These models capture complex linguistic patterns and can generate fluent, coherent summaries that rival human-written abstracts. However, adapting general-purpose language models to specialized domains like medicine introduces practical difficulties. Full fine-tuning of models with hundreds of millions or billions of parameters requires high-memory GPUs, extended training times, and considerable energy consumption [3]. For many research groups and healthcare institutions operating under budget constraints, these requirements present significant barriers to adoption.

Parameter-efficient fine-tuning (PEFT) methods have emerged as an attractive solution to these resource challenges [4, 5]. Rather than updating all model parameters during training, these approaches modify only a carefully selected subset while keeping the majority of pre-trained weights frozen. This strategy dramatically reduces memory requirements during training and enables adaptation on more modest hardware. Despite the growing adoption of these methods in natural language processing, systematic comparisons of their effectiveness for medical summarization remain scarce.

This paper investigates three central research questions: (1) How do parameter-efficient methods such as LoRA and Prompt Tuning compare to full fine-tuning when applied to medical text summarization? (2) What is the relationship between model scale (Small to Large) and summarization performance across different adaptation strategies? (3) What are the optimal trade-offs between parameter efficiency and output quality to inform practical deployment decisions?

Our work makes four primary contributions: (1) we provide a thorough empirical comparison of three adaptation methods across three model scales with statistical analysis over multiple runs; (2) we demonstrate that LoRA consistently outperforms full fine-tuning, challenging conventional assumptions; (3) we conduct sensitivity analyses examining hyperparameter choices; and (4) we distill our findings into actionable guidelines for practitioners deploying medical NLP systems.

## 2. Related Work

Research on automatic summarization of biomedical literature spans several decades and encompasses both extractive and abstractive paradigms [6]. Early systems relied primarily on extractive techniques that identify and concatenate salient sentences from source documents. The advent of neural sequence-to-sequence models enabled abstractive summarization, and Cohan et al. [7] introduced discourse-aware attention mechanisms specifically designed for summarizing long scientific documents, establishing the PubMed dataset as a standard benchmark.

More recently, instruction-tuned language models such as Flan-T5 [8] have pushed performance boundaries further. Van Veen et al. [9] conducted an extensive evaluation showing that large language models adapted to clinical text can match or exceed the performance of human medical experts on summarization tasks. Their findings underscore the potential of these models for real-world clinical applications, though the computational cost of adaptation remains a concern. Recent advances in multi-level representations [10] and cross-lingual approaches [11] have further expanded the toolkit for text understanding tasks.

The high cost of full fine-tuning has motivated research into methods that achieve comparable adaptation with fewer trainable parameters. Low-Rank Adaptation (LoRA), introduced by Hu et al. [4], operates on the hypothesis that weight updates during fine-tuning occupy a low-dimensional subspace. For a pre-trained weight matrix $W \in \mathbb{R}^{d \times k}$, LoRA parameterizes the update as $\Delta W = BA$ where $B \in \mathbb{R}^{d \times r}$ and $A \in \mathbb{R}^{r \times k}$ with rank $r \ll \min(d,k)$, dramatically reducing the number of trainable parameters while maintaining adaptation quality.

Prompt Tuning [5] takes a different approach by prepending learnable continuous vectors to the input sequence while keeping all model parameters frozen. This approach requires extremely few trainable parameters but may struggle to match methods that modify internal representations. Recent work has also explored privacy-preserving approaches to LLM fine-tuning in distributed settings [12], addressing deployment concerns in sensitive domains like healthcare.

Applications of PEFT methods to biomedical NLP have begun to appear in recent literature. Suri et al. [13] applied LoRA to clinical dialogue summarization, while Tang et al. [14] demonstrated effectiveness on medical evidence summarization. Our work extends these investigations by systematically comparing multiple methods across model scales with rigorous statistical analysis.

## 3. METHODOLOGY

### 3.1. Dataset

We conduct experiments on the PubMed Summarization Dataset [7], which pairs full-text scientific articles with their corresponding abstracts. The dataset comprises 119,924 training examples, 6,633 validation examples, and 6,658 test examples. Articles average approximately 3,100 words (standard deviation: 1,847) while abstracts average around 200 words (standard deviation: 83). The corpus contains specialized medical terminology, complex sentence structures, and domain-specific discourse patterns that make summarization particularly challenging.

Input sequences were tokenized using the Flan-T5 tokenizer with a maximum input length of 512 tokens and maximum output length of 150 tokens. Input sequences were prefixed with the instruction "Summarize the following medical article:" to leverage the instruction-following capabilities of Flan-T5. Sequences exceeding the maximum length were truncated from the right.

### 3.2. Models and Adaptation Methods

We evaluate the Flan-T5 model family [8] across three scales: Flan-T5-Small (77M parameters), Flan-T5-Base (248M parameters), and Flan-T5-Large (783M parameters). We selected Flan-T5 for its strong instruction-following capabilities and encoder-decoder architecture well-suited for summarization tasks. We compare three adaptation strategies:

We apply low-rank adaptation matrices to the query (q) and value (v) projection layers in all transformer blocks. Based on preliminary experiments and sensitivity analysis (Section 4.3), we selected rank r=16 and scaling factor α=32, with dropout rate 0.1. This configuration results in approximately 0.6-0.9% trainable parameters depending on model scale.

We prepend 20 learnable soft tokens to the input, initialized from the embedding of the task description text. Only these token embeddings are updated during training, yielding 0.005-0.03% trainable parameters.

We update all model parameters during training, serving as our reference baseline representing the upper bound of adaptation capacity.

Table 1 presents the complete training hyperparameters and configuration details used in our experiments.

Table 1. Complete training hyperparameters and configuration details.

| Parameter | Value | Rationale |
|---|---|---|
| Batch size (per device) | 4 | Memory constraint |
| Gradient accumulation | 4 | Effective batch = 16 |
| Maximum epochs | 10 | Sufficient convergence |
| Early stopping patience | 3 epochs | Prevent overfitting |
| Warmup ratio | 0.1 | Stable optimization |
| Weight decay | 0.01 | Regularization |
| Max input length | 512 tokens | Cover most articles |
| Max output length | 150 tokens | Match abstract length |
| Precision | bfloat16 | Memory efficiency |
| Optimizer | AdamW | Standard for transformers |
| LR (LoRA) | $3\times10^{-4}$ | Grid search optimal |
| LR (Prompt Tuning) | $3\times10^{-2}$ | Higher for few params |
| LR (Full FT) | $5\times10^{-5}$ | Standard for full FT |
| LoRA rank (r) | 16 | Sensitivity analysis |
| LoRA alpha ($\alpha$) | 32 | $\alpha/r = 2$ scaling |
| LoRA target modules | q, v | Following original paper |
| Prompt tokens | 20 | Sensitivity analysis |
| Random seeds | 42, 123, 456 | Statistical robustness |

All experiments were conducted on an NVIDIA RTX A6000 GPU with 48GB memory. Training scripts and configurations are available in our supplementary materials.

### 3.2. Evaluation Metrics

We evaluate using ROUGE scores [15] (ROUGE-1, ROUGE-2, ROUGE-L) to measure n-gram overlap between generated and reference summaries. ROUGE-1 measures unigram overlap reflecting content coverage, ROUGE-2 measures bigram overlap indicating fluency preservation, and ROUGE-L captures longest common subsequence for sentence-level structure. Additionally, we employ BERTScore [16] using DeBERTa-xlarge-mnli to assess semantic similarity beyond lexical matching.

Each configuration was run three times with different random seeds (42, 123, 456). We report mean ± standard deviation and conduct paired t-tests to assess statistical significance of performance differences.

## 4. RESULTS

### 4.1. Main Results

Table 2 presents summarization performance across all combinations of model size and adaptation method. Results show mean ± standard deviation computed over three independent runs with different random seeds.

Table 2. Summarization performance across Flan-T5 model scales and adaptation methods. Best results per model size are achieved by LoRA. All ROUGE scores are F1 percentages.

| Model | Method | Train % | R-1 | R-2 | R-L | BERT-F1 |
|---|---|---|---|---|---|---|
| Flan-T5-Small | Prompt Tuning | 0.027 | 35.56±0.22 | 10.98±0.15 | 19.46±0.18 | 58.76±0.12 |
| Flan-T5-Small | LoRA | 0.886 | 39.04±0.19 | 13.58±0.14 | 22.08±0.16 | 60.87±0.11 |
| Flan-T5-Small | Full FT | 100 | 34.47±0.24 | 11.75±0.18 | 21.02±0.20 | 58.78±0.14 |
| Flan-T5-Base | Prompt Tuning | 0.012 | 36.39±0.21 | 11.72±0.16 | 20.13±0.17 | 59.29±0.13 |
| Flan-T5-Base | LoRA | 0.710 | 41.91±0.17 | 15.93±0.13 | 24.52±0.15 | 62.70±0.10 |
| Flan-T5-Base | Full FT | 100 | 38.65±0.23 | 13.91±0.17 | 22.24±0.19 | 61.00±0.13 |
| Flan-T5-Large | Prompt Tuning | 0.005 | 37.24±0.25 | 12.35±0.18 | 20.89±0.20 | 59.86±0.14 |
| Flan-T5-Large | LoRA | 0.599 | 43.52±0.18 | 17.42±0.14 | 26.18±0.16 | 63.98±0.11 |
| Flan-T5-Large | Full FT | 100 | 40.67±0.21 | 15.63±0.16 | 24.15±0.18 | 62.42±0.12 |

Several patterns emerge from these results. LoRA achieves the highest scores across all metrics and model sizes. On Flan-T5-Large, LoRA obtains 43.52±0.18 ROUGE-1 while updating only 0.6% of parameters, substantially exceeding the 40.67±0.21 achieved by full fine-tuning. The performance gap is statistically significant ($p<0.01$, paired t-test).

The LoRA advantage persists across scales: +4.57 points on Small, +3.26 on Base, and +2.85 on Large compared to full fine-tuning. Prompt Tuning, despite using the fewest parameters, achieves respectable performance, reaching 37.24±0.25 ROUGE-1 on Flan-T5-Large with only 41K trainable parameters. All methods benefit from increased model size, with ROUGE-1 improvements of 4-9 points from Small to Large.

### 4.2. Analysis

Table 3 presents the efficiency comparison for Flan-T5-Large adaptation methods.

Table 3. Efficiency comparison for Flan-T5-Large adaptation methods.

| Metric | LoRA | Full FT | Improvement |
|---|---|---|---|
| Trainable Parameters | 4.7M | 783M | 166× fewer |
| Training Time | 25.4 hrs | 70.9 hrs | 36% of time |
| Peak GPU Memory | 3.12 GB | 12.45 GB | 25% of memory |
| ROUGE-1 Score | 43.52 | 40.67 | +2.85 better |
| Storage per Task | ~18 MB | ~3 GB | 167× smaller |

LoRA requires only 4.7 million trainable parameters compared to 783 million for full fine-tuning (166× reduction), while achieving superior performance. These efficiency gains make LoRA suitable for resource-constrained environments and enable training on consumer-grade GPUs with 8-16GB memory.

To understand the impact of hyperparameter choices and justify our configurations, we conducted sensitivity analyses on Flan-T5-Base.

Table 4 shows LoRA rank sensitivity analysis results averaged over 3 seeds.

Table 4. LoRA rank sensitivity analysis on Flan-T5-Base. Results averaged over 3 seeds.

| Rank (r) | Trainable Params | ROUGE-1 | ROUGE-2 | ROUGE-L |
|---|---|---|---|---|
| 4 | 442K (0.18%) | 40.82±0.21 | 14.87±0.16 | 23.45±0.18 |
| 8 | 884K (0.36%) | 41.34±0.19 | 15.42±0.15 | 24.01±0.17 |
| 16 | 1.77M (0.71%) | 41.91±0.17 | 15.93±0.13 | 24.52±0.15 |
| 32 | 3.54M (1.43%) | 42.05±0.18 | 16.02±0.14 | 24.63±0.16 |
| 64 | 7.08M (2.86%) | 42.12±0.20 | 16.08±0.15 | 24.71±0.17 |

Performance improves substantially from r=4 to r=16, with diminishing returns beyond r=16. The rank r=32 and r=64 configurations show only marginal improvements (+0.14 and +0.21 ROUGE-1) while doubling/quadrupling trainable parameters. We selected r=16 as it provides the best efficiency-performance trade-off.

Table 5 shows Prompt Tuning token count sensitivity analysis results.

Table 5. Prompt Tuning token count sensitivity analysis on Flan-T5-Base. Results averaged over 3 seeds.

| Tokens | Trainable Params | ROUGE-1 | ROUGE-2 | ROUGE-L |
|---|---|---|---|---|
| 10 | 15K (0.006%) | 35.87±0.24 | 11.34±0.18 | 19.78±0.20 |
| 20 | 31K (0.012%) | 36.39±0.21 | 11.72±0.16 | 20.13±0.17 |
| 40 | 61K (0.025%) | 36.21±0.23 | 11.58±0.17 | 19.95±0.19 |
| 80 | 123K (0.050%) | 35.94±0.25 | 11.41±0.19 | 19.82±0.21 |

Performance peaks at 20 tokens and slightly degrades with 40 or 80 tokens, possibly due to increased optimization difficulty or interference with the model's learned input representations. We selected 20 tokens as it provides optimal performance.

## 5. DISCUSSION

Our experiments demonstrate that LoRA not only matches but consistently outperforms full fine-tuning across all model scales. On Flan-T5-Large, LoRA achieves 43.52 ROUGE-1 compared to 40.67 for full fine-tuning—a 2.85 point improvement (p<0.01) while using only 0.6% of trainable parameters. This finding challenges the conventional assumption that more trainable parameters necessarily yield better task adaptation.

All adaptation methods benefit from increased model scale, with ROUGE-1 improvements of 4-9 points from Small to Large. Importantly, LoRA maintains its performance advantage at every scale, suggesting the benefits of low-rank constraints are fundamental rather than scale-dependent. The relative improvement of LoRA over Full FT is actually largest on Small (+4.57) and smallest on Large (+2.85), indicating that smaller models benefit more from the regularization effect.

The optimal choice depends on deployment constraints. For maximum quality with sufficient GPU memory (24+ GB), LoRA on Flan-T5-Large achieves 43.52 ROUGE-1. For limited GPU environments (8-16 GB), LoRA on Flan-T5-Base provides 41.91 ROUGE-1 with 1.8M parameters. For extreme storage constraints, Prompt Tuning offers ~160KB per task with 35-37 ROUGE-1.

The finding that LoRA outperforms full fine-tuning despite using far fewer parameters invites explanation. We propose three complementary mechanisms. First, the low-rank constraint functions as implicit regularization [17], preventing the model from memorizing idiosyncratic patterns in the training data that would not generalize to the test set. Second, LoRA preserves pre-trained representations by keeping original weights frozen, avoiding catastrophic forgetting of general linguistic knowledge acquired during pre-training. Third, the smaller parameter space may yield smoother loss landscapes and more stable gradient updates, leading to better final solutions.

Our experiments were conducted with three random seeds per configuration, providing confidence intervals and enabling statistical significance testing. The consistent patterns observed across all model scales strengthen the validity of our findings.

Several limitations affect generalizability. First, our evaluation uses PubMed abstracts, which may not fully represent other medical text types such as clinical notes, radiology reports, or patient communications. Second, results are specific to the Flan-T5 architecture; other model families (e.g., LLaMA, GPT, Mistral) may exhibit different patterns with PEFT methods. Third, we rely on automatic metrics (ROUGE, BERTScore); human evaluation by clinical experts would provide complementary insights into summary quality and clinical utility. Fourth, while PubMed is a standard benchmark, evaluation on multiple datasets would strengthen generalizability claims.

Based on our findings, we offer the following deployment guidelines. For research or production settings with adequate GPU memory, LoRA on Flan-T5-Large with r=16 and α=32 provides the best quality (43-44 ROUGE-1) with approximately 24 GB GPU memory. For limited GPU environments, LoRA on Flan-T5-Base achieves 41-42 ROUGE-1 with only 8 GB memory. For multi-task deployment or edge scenarios with minimal storage requirements, Prompt Tuning offers 35-37 ROUGE-1 with approximately 160KB per task.

## 6. CONCLUSIONS

This paper presented a systematic comparison of parameter-efficient fine-tuning methods for medical text summarization. Evaluating LoRA, Prompt Tuning, and full fine-tuning across three scales of Flan-T5 with statistical analysis over multiple random seeds, we found that LoRA consistently achieves the best performance while updating less than 1% of model parameters. This finding challenges conventional assumptions about task adaptation and suggests that constraints on adaptation capacity can serve as beneficial regularization.

Our sensitivity analyses demonstrate that LoRA with rank 16 provides an optimal trade-off between efficiency and performance, while Prompt Tuning benefits from moderate token counts around 20. The statistical analysis across multiple runs confirms the reliability of these findings.
For practitioners in medical NLP, our findings indicate that LoRA offers the most favorable trade-off between quality and computational cost, enabling deployment of effective summarization systems on modest hardware. Future work should extend comparison to additional medical text types (clinical notes, radiology reports), explore quantized variants such as QLoRA for further efficiency gains, incorporate human evaluation by clinical experts, and evaluate on decoder-only architectures (LLaMA, Mistral).

**ACKNOWLEDGEMENTS**

This research was conducted as part of the Large Language Models course at Telecom SudParis, Institut Polytechnique de Paris. The authors thank the course instructors for their guidance and the university for providing computational resources.

**AUTHORS**

Ulugbek Shernazarov, Rostislav Svitsov, and Bin Shi are graduate students at Telecom SudParis, Institut Polytechnique de Paris, France. Their research interests include natural language processing, large language models, parameter-efficient fine-tuning, and healthcare informatics.